\documentclass[a4paper, 10pt, conference]{ieeeconf}      % Use this line for a4
\usepackage{FG2026}

\FGfinalcopy % *** Uncomment this line for the final submission

\IEEEoverridecommandlockouts                              % This command is only
\usepackage{cite}
\usepackage{amsmath,amssymb,amsfonts}
\usepackage{graphicx}
\usepackage{textcomp}
\usepackage{multirow}
\usepackage{colortbl}
\usepackage{caption}
\usepackage{subcaption}
\usepackage{booktabs}
\usepackage{float}
\usepackage{newfloat}
\usepackage[skins]{tcolorbox}
\usepackage{hyperref}
\usepackage{doi}
\usepackage{moresize}

\newfloat{promptfloat}{!t}{pop}
\newtcolorbox[auto counter]{prompt}[2]{
    coltitle=black,
    label={prompt:#1},
    colback=gray!10,
    colframe=black!0!black,
    fonttitle=\bfseries,
    title={Prompt \thetcbcounter: #2},
    enhanced,
    fonttitle=\scriptsize, fontupper=\scriptsize, fontlower=\scriptsize, left=1mm, 
    right=1mm, 
    top=1mm, 
    bottom=1mm, 
    middle=1mm,
    arc=0pt,
    boxrule=0pt,
    borderline={1pt}{0pt}{dashed},
    minipage boxed title*=-1.95em,
    attach boxed title to bottom center={yshift=2pt, yshift=0pt},
    boxed title style={enhanced, colback=white!55!white,
    boxrule=5pt, frame hidden}
}

\title{\LARGE \bf
Prompt-to-Gesture: Measuring the Capabilities of Image-to-Video Deictic Gesture Generation
}

\author{\parbox{16cm}{\centering
    {\large 
    Hassan Ali$^{1}$, Doreen Jirak$^{2}$, Luca Müller$^{1}$ and Stefan Wermter$^{1}$}\\
    {\normalsize
    $^1$ Knowledge Technology Group, Department of Informatics, University of Hamburg, Germany\\
    $^2$ Behavioral Lab, Department of Product Development, University of Antwerp, Belgium}}
    \thanks{We acknowledge support by the German Research
Foundation (DFG) under project LUMO (551629603), and Horizon Europe MSCA grants SWEET (101168792), TRAIL (101072488), and GREET (101226624).}
}

\usepackage{fancyhdr}

\begin{document}

\makeatletter
\g@addto@macro\@maketitle{
  \captionsetup{type=figure}\setcounter{figure}{0}
  \def\mycolspace{1.2mm}
  \centering
  \includegraphics[trim=0.0 0.0 0.0 0, clip, width=2.05\columnwidth]{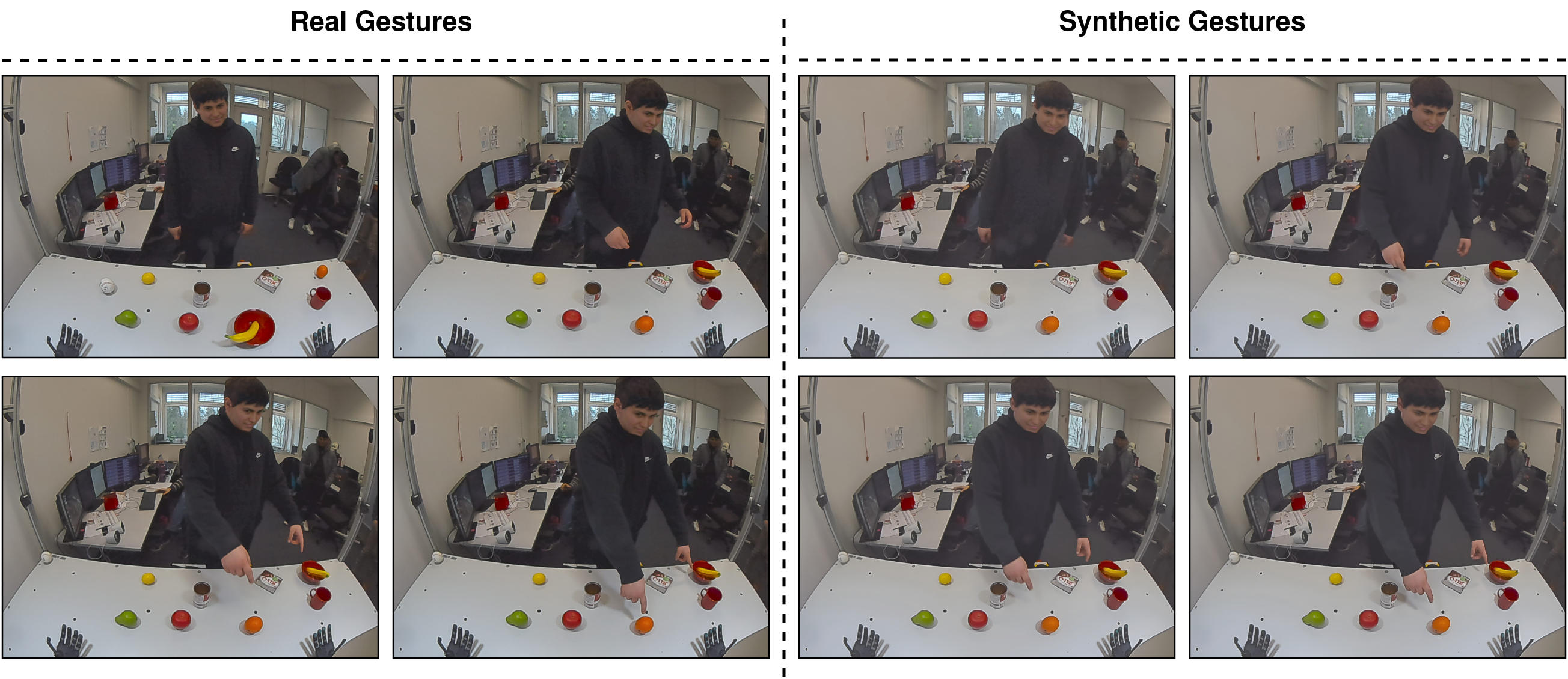}
  \captionof{figure}{Examples of \textbf{real} (left) and \textbf{synthetic} (right) deictic gestures from our dataset, depicting a person pointing to objects in a lab environment. The face of the participant has been replaced with a synthetic one to preserve anonymity.}
    \label{fig:dataset_comparison}
  %\vspace{-3ex}
}

\ifFGfinal
\thispagestyle{empty}
\pagestyle{empty}

\pagestyle{plain}
\fi
\maketitle
\thispagestyle{fancy}
\renewcommand{\headrulewidth}{0pt}
\fancyhf{}
%\fancyhead[C]{2026 International Conference on Automatic Face and Gesture Recognition (FG)}

\fancyfoot[L]{© 2026 IEEE.  Personal use of this material is permitted.  Permission from IEEE must be obtained for all other uses, in any current or future media, including reprinting/republishing this material for advertising or promotional purposes, creating new collective works, for resale or redistribution to servers or lists, or reuse of any copyrighted component of this work in other works.}

%%%%%%%%%%%%%%%%%%%%%%%%%%%%%%%%%%%%%%%%%%%%%%%%%%%%%%%%%%%%%%%%%%%%%%%%%%%%%%%%
\begin{abstract}
Gesture recognition research, unlike NLP, continues to face acute data scarcity, with progress constrained by the need for costly human recordings or image processing approaches that cannot generate authentic variability in the gestures themselves.
Recent advancements in image-to-video foundation models have enabled the generation of photorealistic, semantically rich videos guided by natural language. These capabilities open up new possibilities for creating effort-free synthetic data, raising the critical question of \textit{whether video Generative AI models can augment and complement traditional human-generated gesture data}. In this paper, we introduce and analyze prompt-based video generation to construct a realistic deictic gestures dataset\footnote[3]{\footnotesize{Project Webpage: \url{https://prompt-to-gesture.github.io/}}} and rigorously evaluate its effectiveness for downstream tasks. We propose a data generation pipeline that produces deictic gestures from a small number of reference samples collected from human participants, providing an accessible approach that can be leveraged both within and beyond the machine learning community. Our results demonstrate that the synthetic gestures not only align closely with real ones in terms of visual fidelity but also introduce meaningful variability and novelty that enrich the original data, further supported by superior performance of various deep models using a mixed dataset. These findings highlight that image-to-video techniques, even in their early stages, offer a powerful zero-shot approach to gesture synthesis with clear benefits for downstream tasks.
\end{abstract}

%%%%%%%%%%%%%%%%%%%%%%%%%%%%%%%%%%%%%%%%%%%%%%%%%%%%%%%%%%%%%%%%%%%%%%%%%%%%%%%%
\section{INTRODUCTION}

Gestures are a fundamental component of nonverbal communication. Emerging prior to spoken language~\cite{Corba09,Dick12}, they convey meaning in a wide variety of contexts: pointing gestures that direct attention to objects, symbolic postures such as the ``ok'' sign (in Western culture), beat gestures aligned with speech rhythm, and fully developed systems such as sign languages that can substitute for speech altogether. Within human–robot interaction, research has largely focused on two complementary aspects: the recognition of different gesture classes~\cite{Peral22,Uraka23}, and the generation of gestures to achieve more natural, human-like behavior that avoids uncanny valley effects. Together, these strands of work emphasize the importance of gestures in building intuitive and effective human-machine or human-robot interfaces.

Nevertheless, when considered in the broader context of artificial intelligence research, gestures remain underrepresented. In contrast, natural language processing (NLP) dominates the field, supported by abundant datasets, standardized benchmarks, and an order of magnitude greater publication output. This imbalance has been repeatedly noted in recent surveys and reviews~\cite{khan2025,linardakis2025}, which highlight that the lack of comparable infrastructure has slowed progress in computational models of nonverbal communication. The scarcity of high-quality datasets presents a particular challenge: assembling large-scale gesture corpora requires extensive annotation effort, incurs high financial and time costs, and often results in data that reflect constrained laboratory conditions rather than natural human behavior~\cite{qi2024,jung_2024_crowdsourcing}.

Recent advances in large-scale foundation image-to-video models, such as SORA~\cite{sora_2024_videoworldsimulators2024} and Veo~\cite{veo_2024}, open a promising but, as yet, underexplored avenue to address these challenges. Unlike earlier gesture synthesis approaches, these Generative AI models can produce photorealistic, dynamic human gestures embedded in full scenes, conditioned on natural language and often in a zero-shot manner~\cite{li_2025_zerohsi}. By combining reference imagery with text-based prompts, they enable systematic variation in gesture motion, environmental context, and camera perspective, important conditions that are otherwise cumbersome to obtain in real-world data collection~\cite{gao_2024_challenges}. However, it remains an open scientific question whether AI-generated synthetic gestures achieve comparable fidelity and plausibility to real human recorded data, and whether they can be effectively used to train state-of-the-art gesture recognition models like Transformers.

In this work, we examine the potential of Generative AI models to mitigate the persistent scarcity of large-scale gesture datasets. Specifically, we investigate the ability of state-of-the-art video generation systems to synthesize realistic video sequences of humans performing pointing gestures in a human-robot interaction setup. The contributions of this paper are threefold. First, we provide a computational pipeline for synthesizing gestures using image-to-video models in a zero-shot way to systematically generate and study image-to-video generative models for gesture data synthesis. This pipeline is made accessible and thus easy to use for cross-disciplinary research, e.g., behavioral psychology and robotics. Second, we propose an evaluation framework using quantitative metrics to analyze the realism, diversity, and utility of generated pointing gestures. Third, we employ various machine learning models for gesture classification that demonstrate the feasibility of our approach, which lays the foundation for future research and open challenges at the intersection of gesture understanding and Generative AI.

\begin{figure}
%\vspace{5pt}
\centerline{\includegraphics[width=0.95\columnwidth]{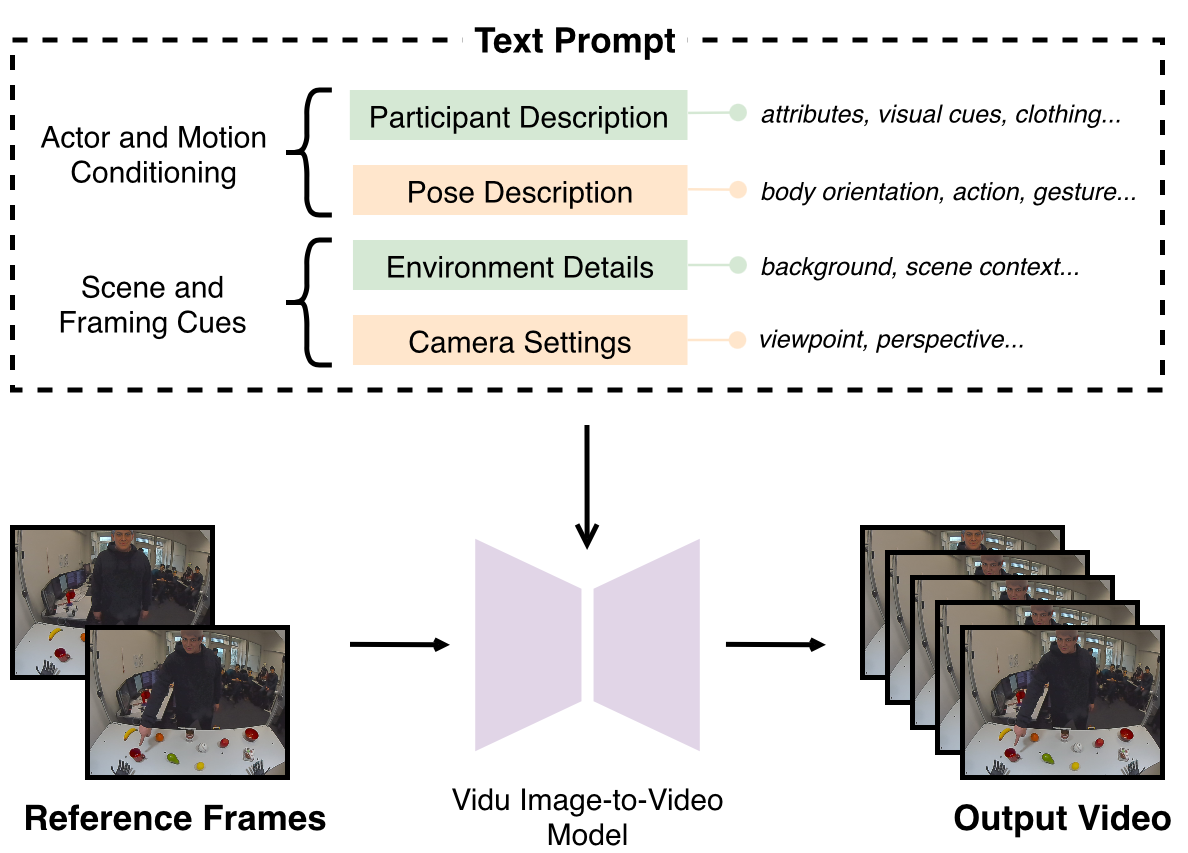}}
\caption{Our synthetic deictic gesture generation pipeline consisting of reference frames and a structured text prompt as input to the Vidu~\cite{bao2024} image-to-video model.}
\label{fig:data_generation_pipeline}
\vspace{-3ex}
\end{figure}

%%%%%%%%%%%%%%%%%%%%%%%%%%%%%%%%%%%%%%%%%%%%%%%%%%%%%%%%%%%%%%%%%%%%%%%%%%%%%%%%
\section{RELATED WORK} 
Recent reviews repeatedly highlight that gesture corpora are often limited in scale, constrained to lab settings and lack gesture variability, which challenges their generalization to ``in-the-wild" and diverse interaction scenarios~\cite{foteinos2025,hashi2024}. Furthermore, gestures encompass a wide range of forms, from simple beats to complex sign language, which further restricts the availability of data for any specific gesture type. For example, the Freihand~\cite{zimmermann_2019_freihand} and H2O~\cite{kwon_2021_h2o} datasets are widely used for hand pose estimation and are therefore particularly suited to (static) hand posture recognition, such as emblems or manipulation gestures like ``picking". Other larger-scale gesture corpora like ChaLearn~\cite{wan_2022_chalearn} (co-speech gestures), IsoGD and ConGD introduce subject variability by a significant number of participants but are recorded in controlled conditions with carefully segmented gestures, besides the costs for recruitment, recording, and labeling. Deictic gestures, most notably \textit{pointing}, play a central role in our research but have thus far been explored mainly in the context of human-machine interaction for gaming in virtual and augmented reality~\cite{Huang16, Das21}. In HRI, research on pointing remains scarce~\cite{Jirak21, Pozzi22}, and existing work is limited to small, scenario-specific corpora. To tackle this gap, work like RenderIH~\cite{li_2023_renderih} shows that synthetic gestures are useful tools for mitigating existing limitations in gesture datasets related to variability in hand pose and the surrounding environment.

%%%%%%%%%%%%%%%%%%%%%%%%%%%%%%%%%%%%%%%%%%%%%%%%%%%%%%%%%%%%%%%%%%%%%%%%%%%%%%%%
Therefore, generating human gestures that are plausible, semantically sound and useful for downstream interaction tasks has become a focused subfield in the gesture community, driving contributions like the GENEA challenge~\cite{Kuche23, Kuche24}. While earlier work relied on recurrent models like LSTMs and Seq2Seq architectures, recent methods increasingly adopt diffusion-based and masked modeling strategies to enhance the diversity and temporal coherence of generated gestures. Models like DiffuGesture~\cite{zhao_2023} and EMAGE~\cite{liu_2024_emage} show such techniques can improve the naturalness and controllability of generated gestures over earlier work~\cite{deichler_2023}. However, current work tends to rely on pose-level representations and does not produce photorealistic videos, limiting applicability to downstream recognition and perception tasks.

\begin{promptfloat}
\vspace{5pt}
\begin{prompt}{prompt_template}{Our gesture generation prompt template with brief examples of the main prompt structure blocks (participant, pose, environment and camera description.)}

\textbf{Participant Description:} \\
The video shows a person in a black hoodie with a logo, standing before a table with items, including a red cup.
\\\\
\textbf{Pose Description:} \\
The person is intently pointing at the \textit{object\_name}. They maintain this pose, drawing attention to the \textit{object\_name} as the main focus. Their posture suggests the \textit{object\_name} is significant, and they remain still throughout the video. The person does not point at any other objects.
\\\\
\textbf{Environment Description:} \\
The scene is set in a modern office. In the background, computer monitors display data, and a group of people is seated in the distance, talking. The lighting is bright and even, creating a professional environment.
\\\\
\textbf{Camera Settings:} \\
Style: Realistic\\
Shot Size: Medium Shot\\
Camera Angle: Eye Level\\
Camera Movement: Static Shot\\
Motion Level: Middle
\end{prompt}
\vspace{-6ex}
\end{promptfloat}

%%%%%%%%%%%%%%%%%%%%%%%%%%%%%%%%%%%%%%%%%%%%%%%%%%%%%%%%%%%%%%%%%%%%%%%%%%%%%%%%
Here, large-scale diffusion-based text-to-video models such as Make-A-Video~\cite{singer_2023_makeavideo} and Imagen~\cite{ho_2022_imagenvideo} come into play, introducing the possibility of high-fidelity video synthesis conditioned on natural language. In addition, image-to-video models like Tune-A-Video~\cite{wu_2023_tuneavideo} and Text2Video-Zero~\cite{khachatryan_2023_text2videozero} allow for one-shot creating and editing of videos based on reference imagery while minimizing the need for model training. Collectively, these advances enable generating videos that are temporally coherent with respect to poses, appearance and surrounding scenery, making them beneficial when exploited for generating full-scene gesture data. AI-generated videos guided by VLMs have been shown to be robust and as effective as real demonstrations for supervised robot learning tasks~\cite{patel_2025}. Other work shows that video generation conditioned on gesture language, like the deictic format, aligns with user intent and is effective for downstream robot imitation learning~\cite{Wang25}. However, image-to-video gesture synthesis as a scalable data source for downstream gesture tasks remains underexplored.

%%%%%%%%%%%%%%%%%%%%%%%%%%%%%%%%%%%%%%%%%%%%%%%%%%%%%%%%%%%%%%%%%%%%%%%%%%%%%%%%
\section{Synthetic Gesture Data Generation} 
In this section, we introduce the original dataset of deictic hand gestures (here: pointing) recorded with human participants and used as a reference during the data synthesis process. Also, we present our synthetic gesture data generation pipeline utilized to augment the deictic gestures dataset. 
\subsection{Real Dataset: Deictic Gestures for HRI}
The dataset was collected using NICOL~\cite{kerzel2023_nicol}, which is a humanoid robot fixed to a table and equipped with two 4K cameras in its left and right eyes. In the recorded scenes, eight participants interacted with the robot by pointing at standard YCB objects, randomly distributed on the table's surface. Each participant pointed to seven random objects as given by the robot's verbal commands. The final dataset consists of 68 short videos, recorded over several hours. The dataset collection process required significant human effort, with two experimenters providing the necessary labor to oversee the data recording process. In addition to the technical setup of the robot, each data recording session consisted of an experiment introduction, consent form signing and questionnaire filling. At the end of each recording session, the robot was reset to its original head position and the objects were repositioned. Since the NICOL robot is based on the Robot Operating System (ROS), a post-processing step was also necessary to extract the collected RGB data from the recorded \textit{.bag} files. The dataset collection process was approved by the Ethics Committee of the Department of Informatics at the University of Hamburg, ensuring all procedures adhered to ethical standards and guidelines. Further, the faces of the participants have been replaced with synthetic faces as an additional measure to protect their identity. Although the real dataset was collected prior to this work, it will be released alongside the synthetic dataset, which is a novel contribution of this paper.

%%%%%%%%%%%%%%%%%%%%%%%%%%%%%%%%%%%%%%%%%%%%%%%%%%%%%%%%%%%%%%%%%%%%%%%%%%%%%%%%
\subsection{Vidu: Image-to-Video Generation}
We employ Vidu~\cite{bao2024}, a state-of-the-art transformer-based image-to-video generation model. Specifically, we use the latest architecture version, \textit{Vidu Q3}, which is based on a U-ViT backbone combined with diffusion-based denoising blocks. We chose this model for various considerations. First, Vidu allows reference-guided generation, which maintains the consistency of the visual presentation of the human participant's appearance and surrounding environment across the the generated video samples. Our preliminary experiments show that this level of visual consistency is still significantly challenging for state-of-the-art models like SORA and Veo. Recent work has identified several shortcomings of mainstream generative models like SORA and Veo, e.g., limitations in complex actions and subtle human features~\cite{liu_2024_sora_review}, misalignments with human preferences~\cite{dai_2024_safesora}, proneness to unreliable and implausible content~\cite{hai_2025_sora_limitation} and inaccuracies in modeling of real-world physics~\cite{motamed_2025_generativevideomodelsunderstand}. In contrast, Vidu exhibits properties that align with the requirements of our gesture generation task, particularly in preserving the visual continuity of the gesturing person and their environment and closely resembling the real-world.

Furthermore, Vidu provides fine-grained control over the generated sequences using natural language, like introducing background noise and varying limb motion speeds, which allows to systematically create diverse external factors that mimic in-the-wild conditions of real-world datasets. Our experiments show that the generated synthetic videos exhibit strong temporal coherence with body dynamics like limb positions that are preserved throughout the frame sequences. Thus, the generation model maintains smooth and natural motion trajectories while minimizing common artifacts like motion jitter and unnatural limb distortions, which are prominent in existing gesture generation pipelines~\cite{liu2025gesturelsmlatentshortcutbased}. 

%%%%%%%%%%%%%%%%%%%%%%%%%%%%%%%%%%%%%%%%%%%%%%%%%%%%%%%%%%%%%%%%%%%%%%%%%%%%%%%%
\subsection{Data Generation Pipeline}
Our data generation pipeline is shown in Figure~\ref{fig:data_generation_pipeline}. To generate synthetic gesture videos, we use reference keyframes sampled from the real dataset. These reference frames act as key points, which guide the model to generate videos that are aligned with the labeled data and annotated gestures of the original dataset. Specifically, we provide the model with both start and end reference frames. The start frame initializes the video sequence, while the end reference frame determines the intended outcome. This data generation pipeline allows the model to create synthetic gestures that preserve the semantics of the task, i.e., pointing to the correct objects on the table, while maintaining the visual and temporal consistency.

We use text-based prompts to guide the model in a zero-shot way. The prompts support the model in preserving the visual integrity of the original scene by providing key cues about the gesturing participant, their hand pose, and the surrounding environment. Additionally, the prompts allow fine-grained control over various aspects of the generated videos like camera viewpoints, gesturing hand motion and background noise. For each prompt, we generate four video samples, each of eight seconds in duration. In total, our synthetic deictic gesture dataset includes 1632 videos. Our prompts are structured into four main parts (see Prompt~\ref{prompt:prompt_template}):

\begin{table}[!t]
\centering
\caption{Comparison between the synthetic and real datasets in terms of hand confidence scores, visual fidelity (FID, FVD) and semantic consistency (CLIP similarity)}
\label{table:pose_conf}
\begin{tabular}{@{}ccccc c@{}}
\toprule
\multirow{1}{*}{\rotatebox[origin=c]{90}{\makebox[0pt][c]{\textbf{Type}}}}                    & \textbf{Dataset}       & \textbf{\begin{tabular}[c]{@{}c@{}}Hand\\Confidence\end{tabular}} & \textbf{FID} & \textbf{FVD} & \textbf{\begin{tabular}[c]{@{}c@{}}CLIP\\Similarity\end{tabular}} \\ \midrule
\multirow{1}{*}{\rotatebox[origin=c]{90}{\textbf{Real}}} 
& \multicolumn{1}{c}{\textbf{Ref}\rule[-0.4cm]{0pt}{0.8cm}} 
& \multicolumn{1}{c}{0.75} 
%& \multicolumn{1}{c}{0.95} 
& \multicolumn{1}{c}{-} 
& \multicolumn{1}{c}{-} 
& \multicolumn{1}{c}{-} \\
\multirow{5}{*}{\rotatebox[origin=c]{90}{\textbf{Synthetic}}} & 
\textbf{Static Scene}  
& 0.77 
%& 0.96                    
& 27.99       
& 25.19      
& 0.95                    
\\
& \textbf{Noisy Scene}   
& 0.77
%& 0.96                    
& 41.13       
& 104.14      
& 0.95                    
\\
& \textbf{Fast Motion}   
& 0.78
%& 0.96                    
& 32.67       
& 54.72       
& 0.96                    
\\
& \textbf{Slow Motion} 
& 0.77
%& 0.96                    
& 28.07       
& 24.02       
& 0.95                    
\\
& \textbf{Dynamic Shift} & 
0.73
%0.95                    
& 65.68       
& 46.96       
& 0.94     
\\
& \textbf{Color Shift} 
& 0.76
%& 0.96                    
& 33.12       
& 20.84       
& 0.96   
\\ \bottomrule
\end{tabular}
\vspace{-1ex}
\end{table}

\begin{enumerate}
\item Participant Description: It provides information about the participant's appearance, like their clothing, to ensure consistency of the participant's visual appearance and identity across the frame sequences.

\item Pose Description: It describes the pointing task and participant's hand motion configurations. This part guides the model in generating synthetic data that aligns with the true labels of the original data. 

\item Environment Description: It outlines the attributes of the surrounding environment. We use this to introduce noise in the scene background like people walking by or moving in the back.

\item Camera Settings: It describes how the scene is visually captured and denotes any specific camera or framing techniques such as filming style, shot size, camera angle and movement. We use these as additional parameters to synthesize hand gesture videos that resemble the original in terms of camera settings but also introduce various augmentations like slow-mo and fast-forward for slower and faster hand motions, respectively.
\end{enumerate}
The videos and prompts are accessible to the gesture research community \footnote[4]{\footnotesize{\url{https://prompt-to-gesture.github.io/}}}, which allows further enhancement of prompt engineering and application of gestures in different, cross-disciplinary settings such as in VR environments, conversational agent design, and behavioral psychology.

%%%%%%%%%%%%%%%%%%%%%%%%%%%%%%%%%%%%%%%%%%%%%%%%%%%%%%%%%%%%%%%%%%%%%%%%%%%%%%%%
\section{Evaluation Methodology of Synthetically Generated Gestures}
Building on the description of the gesture generation pipeline, we now turn to its evaluation. A central concern in this context is not only whether the system produces gestures that appear human-like, but also whether it demonstrates sufficient variability and naturalness to avoid repetitiveness or artificial rigidity. Assessing these dimensions gives insight into the extent to which the generated gestures can effectively support naturalistic interaction and communication. It is also essential to enable deep models to learn from the data and generalize to real human behavior in diverse scenarios.

We inspect the quality of human hand poses in our datasets using hand confidence values derived from MediaPipe, which extracts 21 landmarks for each detected hand, capturing the spatial positions of hand joints (cf. Figure \ref{fig:mp_landmarks}). We use the detection score from MediaPipe's palm detector as an approximation for hand confidence. Since these values are not given in MediaPipe by default, we modified the internal API to expose them. For this experiment, we configure MediaPipe in video mode and set the maximum number of detected hands to two with a 0.5 threshold to mitigate outlier detections. Typically, confidence scores are in the range [0,1] and indicate the likelihood of a detected region corresponding to a real hand. Thus, this provides a reliability measurement, reflecting the accuracy of the predicted hand landmarks, especially considering the scores of the real dataset as a comparison baseline. We collect the scores of the detected hands across the RGB frames and compute the average detection scores. Finally, we compare the hand confidence values of the synthetic gesture data against the real gesture data to determine how accurately the generated hand poses replicate the characteristics of real human poses. Table~\ref{table:pose_conf} reports the hand confidence scores for real and generated gestures across all conditions. The real gestures yield an average score of 0.75, indicating reliable detection despite occasional failures due to rapid pointing motions, occlusions by clothing, or color similarity with background objects. The generated gestures show slightly higher scores in most conditions, except for the ``dynamic shift'', but remain within a comparable range. The results suggest that the generation process does not introduce detection artifacts and reflects natural challenges observed in the reference dataset.

\begin{table}[!t] 
\centering 
\caption{Mean and standard deviation of the motion derivatives (velocity, acceleration and jerk) of the hand motions across the synthetic and real datasets} 
\label{table:motion_derivatives} \begin{tabular}{@{}cccccc@{}} \toprule \multirow{1}{*}{\rotatebox[origin=c]{90}{\makebox[0pt][c]{\textbf{Type}}}} & \textbf{Dataset} & \textbf{\begin{tabular}[c]{@{}c@{}}Velocity\\(1\textsuperscript{st} deriv.)\end{tabular}} & \textbf{\begin{tabular}[c] {@{}c@{}}Acceleration\\(2\textsuperscript{nd} deriv.) \end{tabular}} 
& \textbf{\begin{tabular}[c] {@{}c@{}}Jerk\\(3\textsuperscript{rd} deriv.) \end{tabular}} \\ \midrule \multirow{1}{*}{\rotatebox[origin=c]{90}{\textbf{Real}}} 
& \multicolumn{1}{c}{\textbf{Ref}\rule[-0.4cm]{0pt}{0.8cm}} 
& \multicolumn{1}{c}{0.40 (0.54)} 
& \multicolumn{1}{c}{7.10 (9.85)} 
& \multicolumn{1}{c}{142.70 (197.03)} \\ \multirow{5}{*}{\rotatebox[origin=c]{90}{\textbf{Synthetic}}} 
& \textbf{Static Scene} 
& 0.33 (0.40) 
& 5.55 (6.89) 
& 110.57 (137.07) \\ 
& \textbf{Noisy Scene} 
& 0.47 (0.59) 
& 8.16 (10.63) 
& 165.47 (218.89) \\ 
& \textbf{Fast Motion} 
& 0.44 (0.56) 
& 7.48 (9.71) 
& 149.86 (195.77) \\ 
& \textbf{Slow Motion} 
& 0.32 (0.38) 
& 5.38 (6.57) 
& 106.68 (128.36) \\ 
& \textbf{Dynamic Shift} 
& 0.44 (0.50) 
& 7.66 (8.65) 
& 153.53 (170.51) \\ 
& \textbf{Color Shift} 
& 0.41 (0.70) 
& 7.36 (12.83) 
& 138.12 (194.37) \\ 
\bottomrule 
\end{tabular} 
\vspace{-1ex}
\end{table}

\begin{figure*}
    \vspace{5pt}
    \centering
        \begin{subfigure}[b]{0.32\textwidth}
            \centering
            \includegraphics[width=0.95\textwidth]{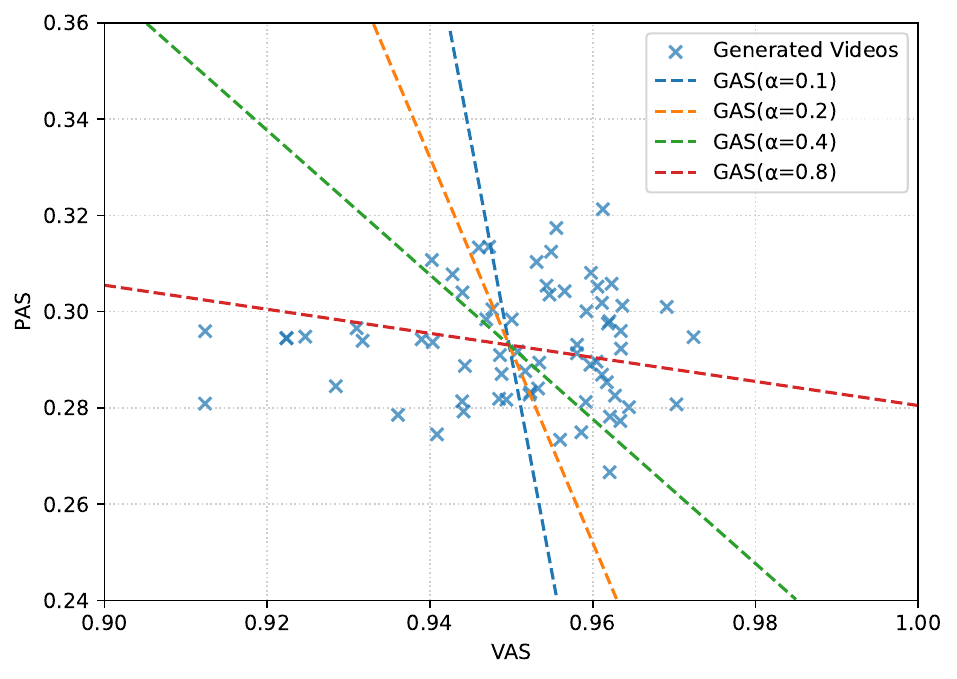}
            \vspace{-1ex}
            \caption{Static Scene}
            %\vspace{3mm}
        \end{subfigure}
        \begin{subfigure}[b]{0.32\textwidth}
            \centering
            \includegraphics[width=0.95\columnwidth]{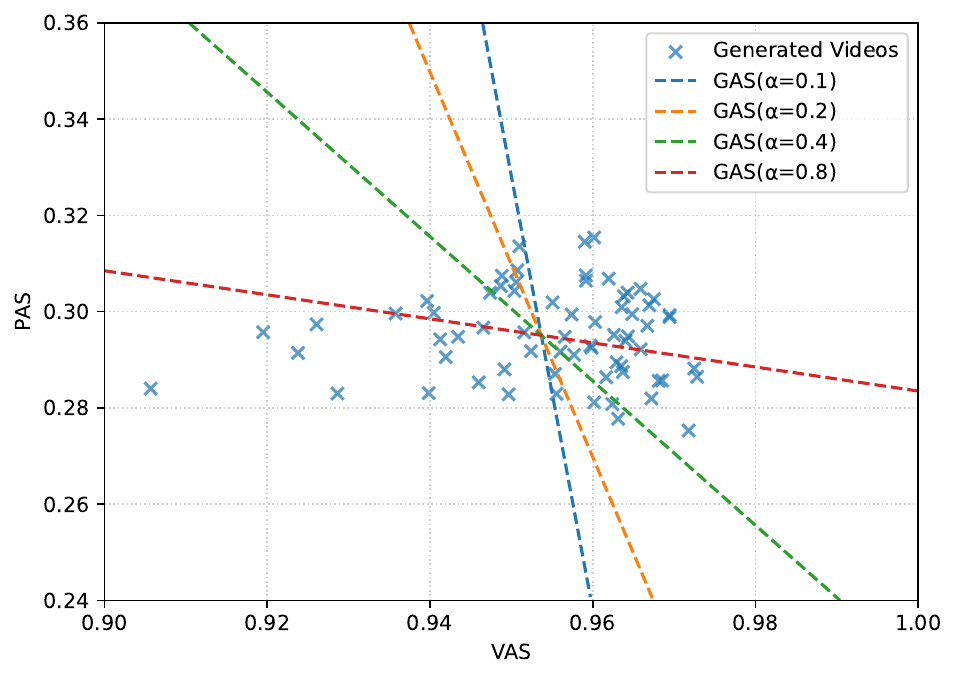}
            \vspace{-1ex}
            \caption{Dynamic Scene}
            %\vspace{3mm}
        \end{subfigure}
        \begin{subfigure}[b]{0.32\textwidth}
            \centering
            \includegraphics[width=0.95\textwidth]{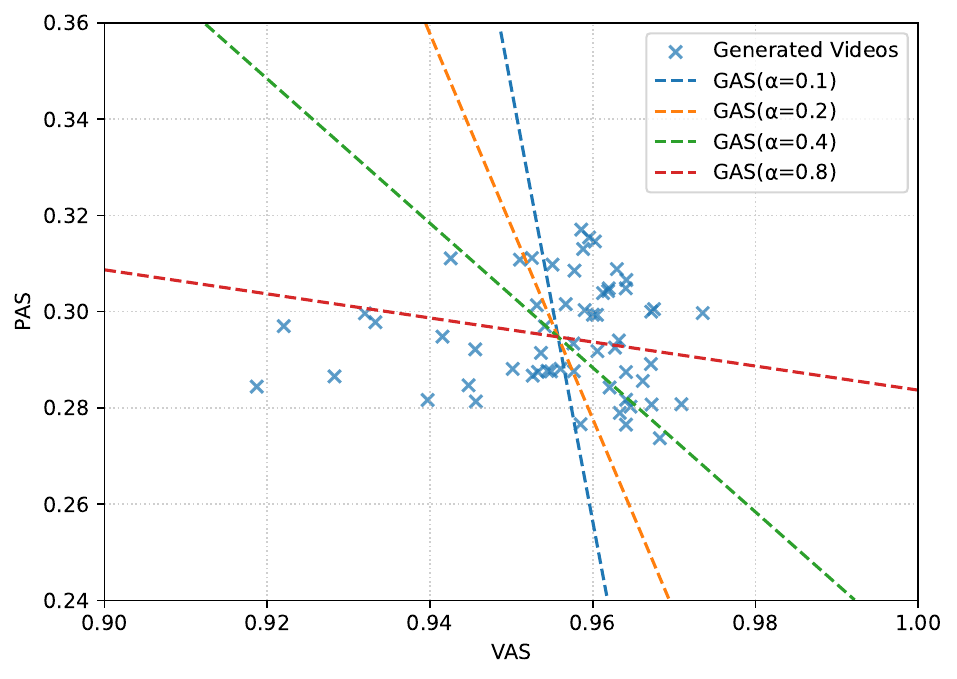}  
            \vspace{-1ex}
            \caption{Fast Motion}
            %\vspace{-9.5pt}
        \end{subfigure}
        \\[3pt]
        \begin{subfigure}[b]{0.32\textwidth}
            \centering
            \includegraphics[width=0.95\textwidth]{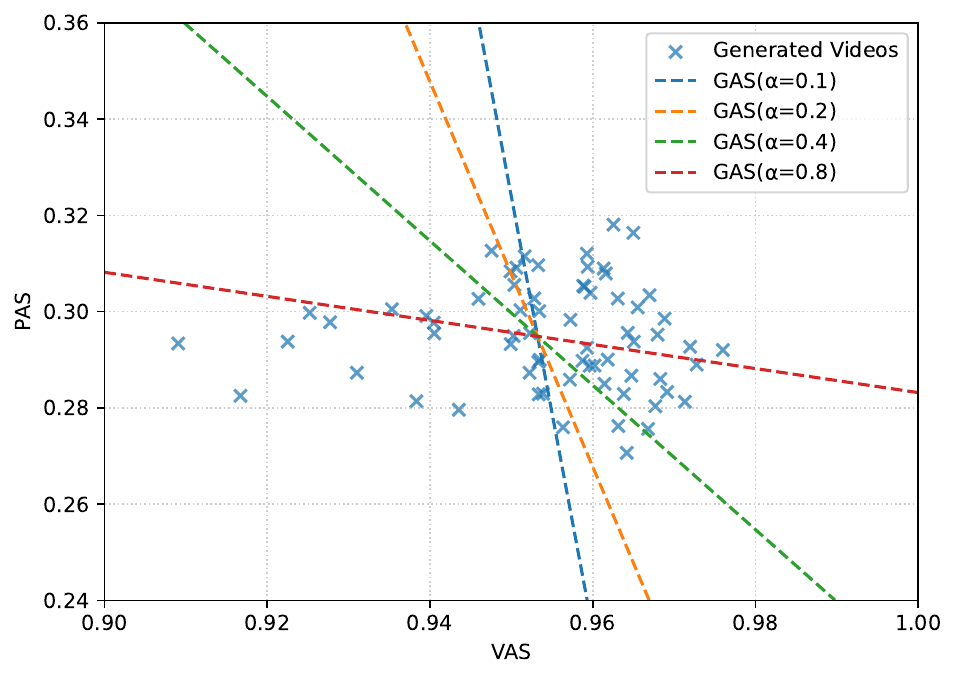}
            \vspace{-1ex}
            \caption{Slow Motion}
            \label{plot_clip_similarity}
            %\vspace{3mm}
        \end{subfigure}
        \begin{subfigure}[b]{0.32\textwidth}
            \centering
            \includegraphics[width=0.95\textwidth]{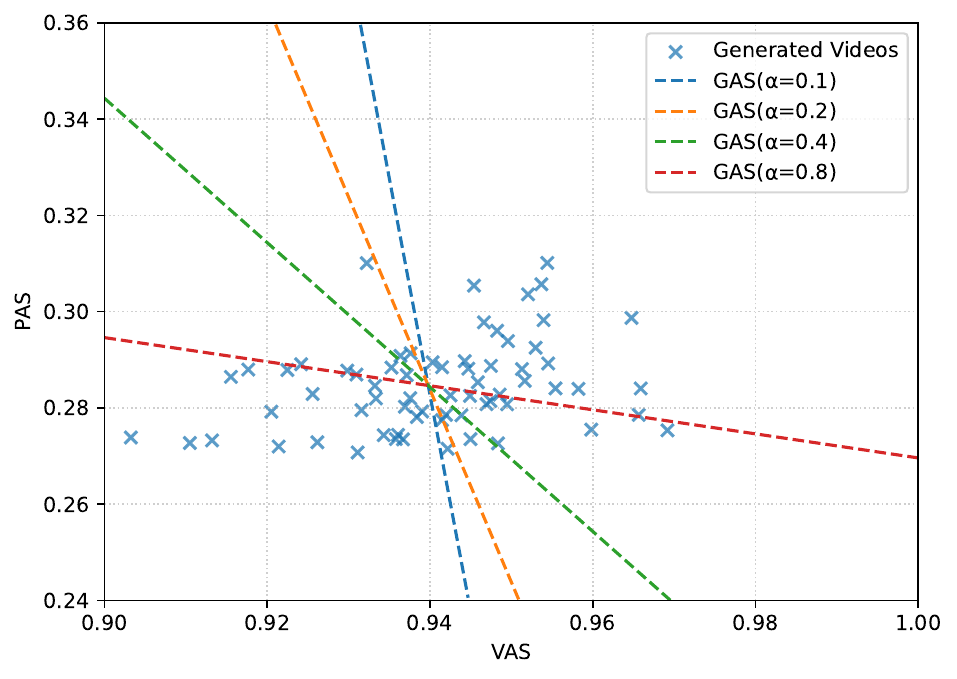}
            \vspace{-1ex}
            \caption{Dynamic Shift}
            \label{plot_clip_similarity}
            %\vspace{3mm}
        \end{subfigure}
        \begin{subfigure}[b]{0.32\textwidth}
            \centering
            \includegraphics[width=0.95\textwidth]{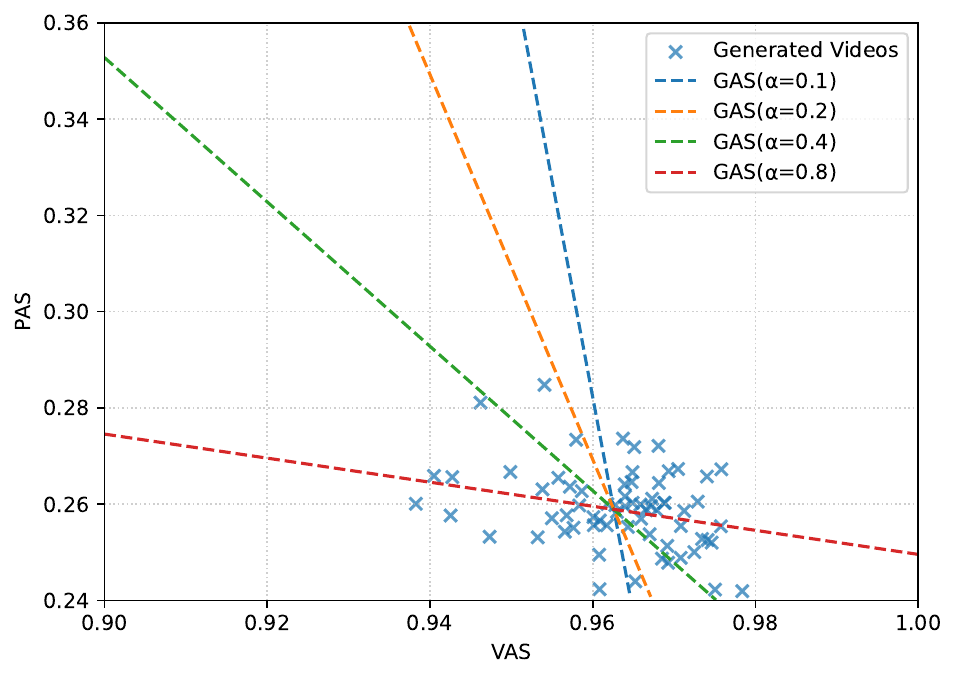}
            \vspace{-1ex}
            \caption{Color Shift}
            \label{plot_clip_similarity}
            %\vspace{3mm}
        \end{subfigure}
        
        \caption{Gesture Alignment Scores (GAS) for different conditions. Each synthetic video is represented by a pair of Visual Alignment Score (VAS) and Prompt Alignment Score (PAS). The dashed lines are iso-GAS lines for different $\alpha$ values.}
    \label{fig:hgs}
    \vspace{-2ex}
\end{figure*}

%%%%%%%%%%%%%%%%%%%%%%%%%%%%%%%%%%%%%%%%%%%%%%%%%%%%%%%%%%%%%%%%%%%%%%%%%%%%%%%%
\subsection{Distribution and Semantic Alignment}
After obtaining a global measure of hand detection confidence, we proceed with fine-grained metrics that quantify the differences between real and generated gesture data distributions. That is, we look into the Fréchet Inception Distance (FID) and Fréchet Video Distance (FVD). While FID compares still images, FVD is used for videos. This way, we can capture different aspects of data fidelity like frame-level visual quality and temporal coherence. Low FID/FVD values indicate close resemblance between real and synthetic gestures where 0 is a perfect alignment. However, there is no upper bound for either metric, unlike confidence scores, but they provide a relative scale that needs careful interpretation depending on the underlying data distribution. To calculate FID, we use a pre-trained Inception-v3 model (as suggested by Heusel et al.~\cite{heusel2017_fid} allowing comparability with prior work) to create image embeddings of both the real and synthetic datasets. For the synthetic videos, we calculate an average over the four samples generated by each text-prompt, thus representing a single data sample. Then, we compute the Fréchet distance between two multivariate Gaussian distributions approximating the original and synthetic datasets. For computational efficiency, we sample every fifth RGB frame.

Similarly, we calculate FVD using a pre-trained 3D CNN (as suggested by Unterthiner et al.~\cite{unterthiner_2019_fvd}), specifically a 3D ResNet-50 loaded with PyTorch, to capture the spatiotemporal video features. 
Table~\ref{table:pose_conf} indicates that the static and slow conditions closely resemble the real data, with both FID and FVD below 30. The color shift condition also resembles the reference data since it focuses only on visual appearance (color and illumination). In contrast, the noisy and fast motion conditions exhibit a pronounced divergence between the metrics: although FID suggests reasonably realistic frames (41.13/32.67), FVD is substantially higher (104.14/54.72), highlighting deficiencies in temporal coherence. The results illustrate a common challenge in video generation: while generating plausible individual frames is relatively straightforward, producing smooth, temporally consistent gestures remains difficult, often leading to jitter or unnatural transitions, for instance, when moving from a pointing gesture to a resting arm position. Conversely, the dynamic shift condition shows the opposite pattern: FVD is comparatively low, demonstrating robustness in capturing temporal characteristics, while FID is significantly higher, indicating the limitations of relying on frame-level quantification alone.

Finally, we inspect similarity scores using Contrastive Language-Image Pre-training (CLIP~\cite{radford2021_clip}), which is a technique that involves image and text encoders widely used to connect visual embeddings with abstract concepts. Since CLIP was trained on millions of images, we use it to measure the semantic alignment between the real and synthetic datasets in terms of content. To calculate CLIP similarity, we sample every fifth frame and extract image embeddings using a ViT-L/14 backbone. We compare the extracted embeddings of the real and synthetic data using cosine similarity. The similarity scores (see Table~\ref{table:pose_conf}) are consistently high across all control factors. Since CLIP is pre-trained on static scenes, the results suggest strong alignment between the datasets in visual content, e.g., a person standing before a table, objects positioned on top, a lab in the background. Thus, this showcases that while the synthetic data introduces noticeable variability (as indicated by FID and FVD), it semantically resembles the original data, preserving meaningful motion patterns and relationships that make it suitable for downstream tasks like gesture recognition or predictive modeling.

%%%%%%%%%%%%%%%%%%%%%%%%%%%%%%%%%%%%%%%%%%%%%%%%%%%%%%%%%%%%%%%%%%%%%%%%%%%%%%%%
\subsection{Gesture Alignment Scores}
%\subsection{Human Gesture Similarity}
We further use CLIP to measure the similarity between the generated and reference datasets across two dimensions. First, we measure the cosine similarity of the image embeddings (as described in the previous section), referred to here as the Visual Alignment Score (VAS). Second, we examine the text prompt alignment by measuring the cosine similarity between the image and text embeddings of the intended gesture description, which we refer to as the Prompt Alignment Score (PAS). It is worth noting that PAS values are relatively low for both the real and synthetic datasets, i.e., even for the reference videos collected with human participants. This can be explained by the fact that CLIP is frame-based and does not create temporal representations. Thus, while it is useful in capturing appearance-based cues and visual information, it struggles with temporal actions that unfold over frame sequences. Consequently, subtle actions and hand motions such as deictic gestures are challenging. In order to obtain more balanced insights into the datasets, we propose a new metric combining VAS and PAS into a single metric, which we call the Gesture Alignment Score (GAS), to assess the videos in terms of semantic fidelity and visual realism in a controllable way. For each video, given the VAS and PAS values, we calculate the gesture alignment score as a linear combination weighted by $\alpha \in [0,1]$:

\begin{figure}
%\vspace{5pt}
\centerline{\includegraphics[width=0.5\columnwidth]{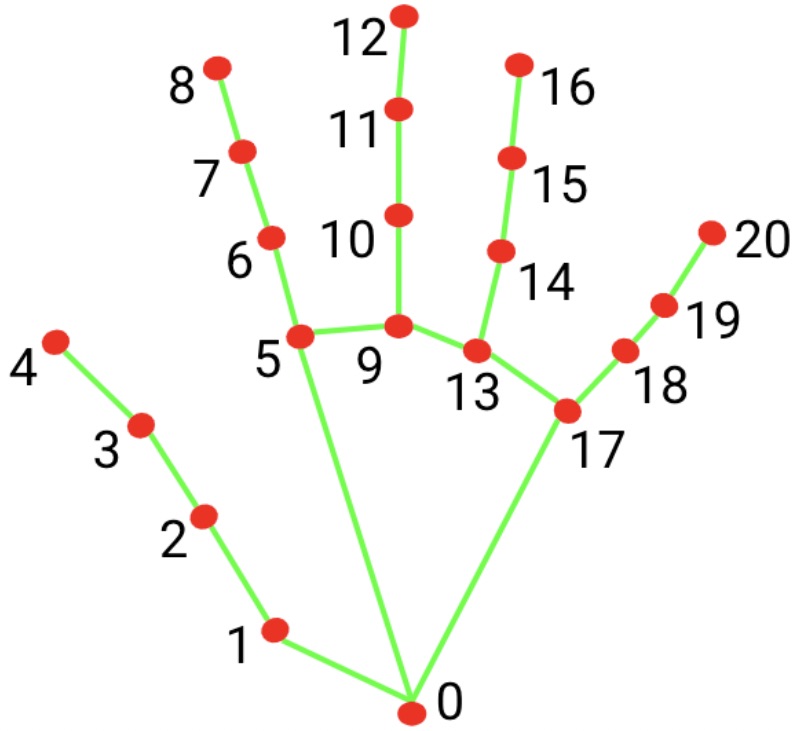}}
\caption{The 21 hand landmarks provided by MediaPipe and corresponding to finger joints and wrist keypoints~\cite{zhang2020mediapipe}.}
\label{fig:mp_landmarks}
\vspace{-3ex}
\end{figure}

\begin{equation}
\text{GAS} = \alpha \cdot \text{PAS} + (1-\alpha) \cdot \text{VAS}
\end{equation}

Figure~\ref{fig:hgs} plots synthetic videos PAS and VAS values under static/noisy scenes, slow/fast motion and camera/color shifts with dashed iso-GAS lines representing mean GAS values for several $\alpha$ levels ($0.1$, $0.2$, $0.4$, $0.8$). Lower $\alpha$ emphasizes visual naturalness, while higher $\alpha$ enforces adherence to text prompts. Most conditions cluster tightly around the iso-GAS lines, aligning with human references, whereas fast motion and camera shift conditions show broader spreads, reflecting higher variability in complex transitions, i.e., hand speed and camera shifts. Slower motions align more closely with human reference videos. Although samples in the color shift condition slightly shift across the PAS, VAS values remain high, indicating consistent visual alignment despite the appearance changes. Also, the clustering behavior around the iso-GAS lines remains consistent, showing that even when one aspect (e.g., PAS) varies, the overall gesture quality as measured by GAS remains stable. This demonstrates that GAS is sufficiently sensitive to detect instability in complex settings (e.g., fast motion and camera shift) while confirming robustness where synthetic gestures remain reliably usable.

%%%%%%%%%%%%%%%%%%%%%%%%%%%%%%%%%%%%%%%%%%%%%%%%%%%%%%%%%%%%%%%%%%%%%%%%%%%%%%%%
\subsection{Motion Derivatives}
Techniques like CLIP focus on the overall content of the videos and do not provide insights into the hand motions, especially for subtle gestures. Therefore, we look into various motion derivatives, i.e., velocity, acceleration and jerk, in order to gain a better understanding of how the hand motion changes over time in the synthetic dataset and in comparison to the real dataset. We calculate the motion derivatives by extracting joint positions of the human hands with MediaPipe. Using the 3D positions of the hand landmarks (Figure \ref{fig:mp_landmarks}), we compute the frame-to-frame changes in their trajectories. Based on the spatial representations, we compute the first derivative (velocity), second derivative (acceleration), and third derivative (jerk) with respect to time. By computing the magnitude of the resulting vectors, we obtain scalar values that capture the motion intensity at each frame (see Table~\ref{table:motion_derivatives}). 

Overall, synthetic gestures exhibit motion profiles similar to the real dataset while reflecting variability from different augmentation conditions. Noisy scenes show higher motion derivatives, static and slow conditions produce subtle motion, dynamic shifting scenes increase activity without disrupting smoothness, while color shifting scenes remain close to the reference videos, showing that the variability lies in the appearance rather than in the motion. These results confirm that video generation models preserve human-like gestures suitable for downstream augmentation.

\begin{table}[!t]
\centering
\caption{KL divergence and EMDs between the synthetic and real datasets based on finger joint angles. The MediaPipe joints represent specific finger landmarks (see Figure~\ref{fig:mp_landmarks})}
\label{table:joint_angles}
\begin{tabular}{cccccc}
\toprule
\textbf{Finger} & \textbf{MP Joint} & \textbf{Type} & \textbf{KL Divergence} & \textbf{EMD} \\
\midrule
\multirow{3}{*}{\textbf{Thumb}}  
& 1 
& CMC 
& 0.05 
& 0.91 
\\
& 2 
& MCP  
& 0.06 
& 1.69 
\\
& 3 
& IP 
& 0.02 
& 1.54 
\\
\midrule
\multirow{3}{*}{\textbf{Index}}  
& 5 
& MCP 
& 0.05 
& 2.31 
\\
& 6 
& PIP 
& 0.15 
& 14.86 
\\
& 7 
& DIP 
& 0.11 
& 4.97 
\\
\midrule
\multirow{3}{*}{\textbf{Middle}} 
& 10 
& MCP 
& 0.03 
& 2.49 
\\
& 11 & PIP & 0.09 & 3.45 \\
& 12 & DIP & 0.08 & 2.61 \\
\midrule
\multirow{3}{*}{\textbf{Ring}}   
& 14  
& MCP 
& 0.06 
& 2.64 
\\
& 15 
& PIP 
& 0.09 
& 2.88 
\\
& 16 
& DIP 
& 0.06 
& 2.29 
\\
\midrule
\multirow{3}{*}{\textbf{Pinky}}  
& 18 
& MCP 
& 0.05 
& 2.63 
\\
& 19 
& PIP 
& 0.07 
& 3.14 
\\
& 20 
& DIP 
& 0.06 
& 1.77 
\\
\bottomrule
\end{tabular}
\vspace{-2ex}
\end{table}

%%%%%%%%%%%%%%%%%%%%%%%%%%%%%%%%%%%%%%%%%%%%%%%%%%%%%%%%%%%%%%%%%%%%%%%%%%%%%%%%
\subsection{Joint Angle Distribution}
While the motion profiles summarize the hand movements, they do not capture the subtle variations of joint angles across frames. Therefore, we look into the Kullback–Leibler (KL) divergence and Earth Mover’s Distance (EMD) between the distributions of angles in the synthetic and real data. We calculate three joint angles of each of the five hand fingers using landmarks from MediaPipe. The three angles correspond to the fingertip, Metacarpophalangeal (MCP), Proximal Interphalangeal (PIP) and Distal Interphalangeal (DIP) joints of the index, middle, ring and pinky fingers. Since the thumb finger has a single joint, we consider the angle at the Carpometacarpal (CMC) joint, i.e., between the wrist and MCP joint, corresponding to MediaPipe hand landmarks numbers 0, 1 and 2 (cf.~Figure~\ref{fig:mp_landmarks}), respectively. 

The joint angles in the synthetic data show moderate KL divergence from the real data (see Table~\ref{table:joint_angles}), indicating well-aligned finger configurations. The synthetic gestures replicate a typical pointing pose: an extended index finger and remaining fingers curled toward the palm. While EMD values for the PIP and DIP joints of the index finger are higher, i.e., reflecting natural variability amplified in the synthetic data, the pipeline accurately reproduces hand posture and anatomical coordinates. In contrast, the MCP joint of the index finger plays the most prominent role in defining the direction of the pointing gesture. Its lower divergence shows that the synthetic dataset effectively preserves the essential structure and semantics of pointing gestures, accurately reproducing key finger configurations, particularly the MCP joint of the index finger, despite natural variability in less critical joints.

\begin{table}[!t]
\caption{Accuracy and F-1 scores of the machine learning models trained on the real and synthetic datasets using three training paradigms (baseline, pre-training and fine-tuning)}
\centering
\label{table:accuracy_results}
\begin{tabular}{@{}lcccccc@{}}
\toprule
& \textbf{Experiment}     
& \textbf{\begin{tabular}[c]{@{}c@{}}Training\\ Data\end{tabular}} 
& \textbf{\begin{tabular}[c]{@{}c@{}}Test\\ Data\end{tabular}} 
& \textbf{Accuracy} 
& \textbf{\begin{tabular}[c]{@{}c@{}}F-1 \\ Score\end{tabular}} 
\\ \midrule

% CNNLSTM
\multirow{3}{*}{\begin{tabular}[c]{@{}c@{}}\rotatebox{90}{\ssmall{CNNLSTM}}\end{tabular}} 
& \textbf{Baseline} & Real & Real & 0.909 & 0.799 \\ 
& \textbf{Pre-training} & Synthetic & Real & 0.887 & 0.848 \\ 
& \textbf{Fine-tuning} & Synthetic+Real & Real & \textbf{0.937} & \textbf{0.918} \\ 
\addlinespace[1ex]\midrule

% MM-ITF
\multirow{3}{*}{\begin{tabular}[c]{@{}c@{}}\rotatebox{90}{\ssmall{MM-ITF}}\end{tabular}} 
& \textbf{Baseline} & Real & Real & 0.891 & 0.892 \\ 
& \textbf{Pre-training} & Synthetic & Real & 0.872 & 0.874 \\ 
& \textbf{Fine-tuning} & Synthetic+Real & Real & \textbf{0.897} & \textbf{0.898} \\ 
\midrule

% VideoMAE
\multirow{3}{*}{\begin{tabular}[c]{@{}c@{}}\rotatebox{90}{\ssmall{VideoMAE}}\end{tabular}} 
& \textbf{Baseline} & Real & Real & 0.850 & 0.780 \\ 
& \textbf{Pre-training} & Synthetic & Real & 0.923 & 0.866 \\ 
& \textbf{Fine-tuning} & Synthetic+Real & Real & \textbf{0.950} & \textbf{0.944} \\ 
%\midrule

\bottomrule
\end{tabular}
\vspace{-2ex}
\end{table}

%%%%%%%%%%%%%%%%%%%%%%%%%%%%%%%%%%%%%%%%%%%%%%%%%%%%%%%%%%%%%%%%%%%%%%%%%%%%%%%%
\subsection{Synthetic Data Diversity}
\begin{figure}[!t]
\centering
%\vspace{5pt}
\centerline{\includegraphics[width=0.95\columnwidth]{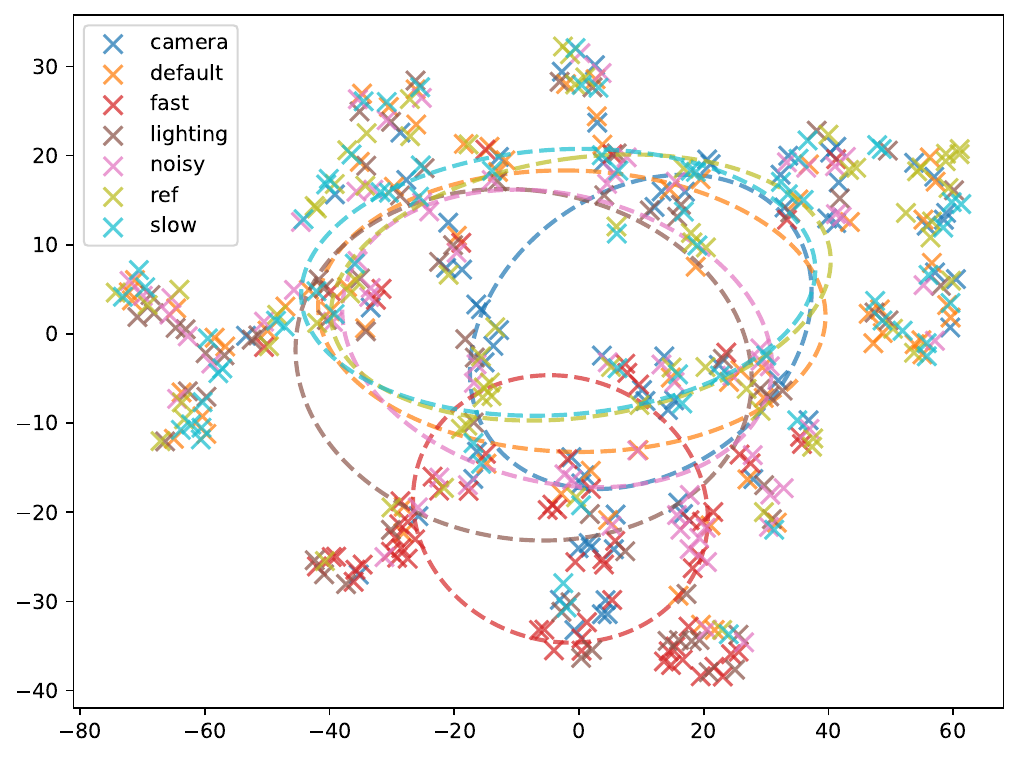}}

\caption{The synthetic and real datasets projected into 2D space using t-SNE. Each point represents an averaged vector representation of 3D hand landmarks over frame sequences.}
\label{fig:diversity_scores}
\vspace{-3ex}
\end{figure}

Several previous metrics showcase variability in the generated synthetic data, indicating the effectiveness of the augmentation text prompts in guiding the data generation pipeline to create several variations of the same data. To extend this analysis, we look into the diversity and richness of the synthetic gestures by analyzing the 3D hand landmark embeddings, thus allowing us to quantify variations in both the inter- and intra-conditions and visualize how well the synthetic gestures reflect the range of hand configurations observed in real hand gestures. We create data representations by stacking 63-dimensional vectors corresponding to 3D hand landmarks from MediaPipe across the RGB frames. After calculating mean video embeddings, we find the inter- and intra-control factor distances using cosine distances. We apply t-SNE (2 components, 10 perplexity) to the distances to visualize diversity across the data augmentation factors in comparison to the reference data. In Figure~\ref{fig:diversity_scores}, we visualize ellipses centered at mean 2D projections of each condition, with the orientation reflecting covariance and size scaled to one standard deviation. The t-SNE projections show strong overlap between synthetic and real data, except for the fast motion condition, which separates due to added dynamics like arm swings or abrupt movements. These variations, while absent in the original data, are plausible and maintain correct pointing behavior. Similar ellipse sizes across conditions indicate the synthetic data preserves variability levels comparable to real participants. Overall, the diversity scores confirm the augmentation pipeline introduces realistic variability while maintaining alignment with real gestures.

%%%%%%%%%%%%%%%%%%%%%%%%%%%%%%%%%%%%%%%%%%%%%%%%%%%%%%%%%%%%%%%%%%%%%%%%%%%%%%%%
\subsection{Synthetic-to-Real Knowledge Transfer in Machine Learning Models}
Assessing the fidelity of synthetic gestures ensures it is not only meaningful for cross-disciplinary researchers employing this data but also provides a reliable foundation for training gesture recognition models, guaranteeing that AI systems learn from accurate, semantically valid motions rather than merely visually plausible videos. Collecting and annotating human data is costly and often limited in scale, while synthetic data can be automatically generated in larger quantities. However, this does not guarantee that the knowledge will transfer across the various data sources. The goal of this experiment is to evaluate: 1) whether the training with our synthetic data can generalize to real human behavior, and 2) whether pre-training on our synthetic data followed by fine-tuning on the real data can boost the overall model performance compared to a baseline trained on real data only. 

In our experiment, we use three models: 1) CNNLSTM with attention, a well-known established recurrent model for gesture recognition, 2) Multi-Modality Inter-Transformer (MM-ITF)~\cite{moeller2025}, previously successfully evaluated on the real dataset referenced in this paper, 3) VideoMAE~\cite{tong_2022_videomae}, a transformer-based masked autoencoder for video representation learning. Collectively, these models evaluate the usability of our synthetic gestures under different modalities (RGB in CNNLSTM and VideoMAE, hand and object poses in MM-ITF) and various applications (spatiotemporal gesture modeling in CNNLSTM, hand-object relationship modeling in MM-ITF and video classification in VideoMAE)

We conducted three experiments to evaluate the models under different training paradigms. The \textit{baseline} used only real data, the \textit{pre-training} used synthetic data tested on real out-of-distribution (OOD) data, and the \textit{fine-tuning} hybrid approach further trained the synthetic-based model with real data. All experiments were evaluated on the same 30\% hold-out real test set (Table~\ref{table:accuracy_results}). Results show that hybrid training outperforms both pure real and synthetic setups across all models, indicating that pre-training on synthetic data provides a strong initialization for downstream applications. Comparable baseline and pre-training performance highlights the utility of synthetic data for generalizing to OOD scenarios, while the added variability in synthetic data supports its value for future training and applications\footnote[5]{\footnotesize{Further information on our model training on project webpage: \url{https://prompt-to-gesture.github.io/}}}.

%%%%%%%%%%%%%%%%%%%%%%%%%%%%%%%%%%%%%%%%%%%%%%%%%%%%%%%%%%%%%%%%%%%%%%%%%%%%%%%%
\section{DISCUSSION AND LIMITATIONS}
Gesture recognition research continues to suffer from data scarcity due to the challenges of recruiting participants and introducing natural variability, which has negative impacts on gesture recognition applications. Most studies that rely on small, application-specific datasets ~\cite{Jirak21} and common augmentation techniques, limited to image manipulations or time scaling, fail to capture realistic gesture variations. Moreover, \textit{gestures} is the overall term that encompasses a wide variety of types, and designing benchmarks for specific categories such as deictic (pointing) gestures is particularly challenging due to variations in pointing behavior and the influence of object arrangements on the pointing direction. Beyond machine learning, gestures are also central in fields such as human-machine interaction ~\cite{Das21}, behavioral psychology~\cite{Kim24}, and robotics~\cite{Pozzi22}, where they enhance communication and inform the study of human perception and action. 
Thanks to the progress in Generative AI for videos ~\cite{bao2024, singer_2023_makeavideo}, we addressed the recent issues in gesture research by introducing a novel gesture generation pipeline, accompanied by exhaustive evaluation techniques that can serve as a guideline for (cross-disciplinary) researchers.

Our quantitative analyses show that our method generates gestures with high fidelity to real reference data, capturing human-like motion across multiple conditions as measured by FID, FVD and physical metrics like velocity derivatives. While challenges in video generation remain, our method introduces realistic variability, mitigates lab-induced biases and provides a trustworthy data source for downstream machine learning and other applications. Moreover, the generated gestures, videos and prompts are accessible to researchers without extensive ML expertise, enabling reproducible deictic gestures across contexts or serving as a shortcut for producing other gesture types. We also transformed our analyses into actionable insights, providing the ML community with validated guidelines and resources for generating and leveraging realistic gesture data. Specifically, our knowledge transfer experiment with several deep models revealed higher accuracy when using the generated data, which supports our claim that the synthetic data is useful for downstream gesture recognition tasks, further showcasing that synthesizing gestures using Generative AI is powerful for gesture research.

However, it should be noted that the use of current Generative AI video models still entails limitations and constraints, which directly impact our work. Our proposed text-prompt structure provides control over various generation conditions in a zero-shot way, which proves effective in introducing various environmental noise and motion speeds. However, the current generation of models lacks in-depth fine-grained control parameters like guidance scale, seed, sampling temperature and noise level, which exist in state-of-the-art diffusion models. These parameters can improve control over the generation outcome, consistency and reproducibility. During the generation process, we observed a tendency and internal bias in the generative model for cinematic effects and anime-like motions, which we hypothesize is due to its training data. Also, we observed some filler motions where the participant raised the arm unexpectedly in the air despite being prompted otherwise as supported by quantitative scores dealing with motions conditions. Such instances of synthetic bias should be taken into consideration when using these tools.
Moreover, while reference frames effectively guide the model during generation, current image-to-video models lack video-based in-context learning for few-shot prompting. The ability to teach the model a small number of regional or culture-specific gestures, currently rarely represented in standard datasets, would greatly enhance the diversity and representativeness of generated gesture data.

%%%%%%%%%%%%%%%%%%%%%%%%%%%%%%%%%%%%%%%%%%%%%%%%%%%%%%%%%%%%%%%%%%%%%%%%%%%%%%%%
\section{CONCLUSION}
We proposed a synthetic data generation pipeline powered by Generative AI, capable of producing photorealistic gesture videos aligned with HRI setups at scale. Our results show that the generated gestures closely resemble real human gestures while introducing novel in-the-wild variability that reflects real-world dynamics. While Generative AI models are still emerging, they hold great potential not only for augmenting small-scale datasets like ours but also for creating entirely new ones. This is valuable for improving the generalization and accuracy of state-of-the-art gesture recognition models in HRI applications, also for underrepresented gestures, culture-dependent gestures and gestures in social settings. Future work will explore these aspects of gesture generation. We will also expand beyond deictic gestures to evaluate our data generation pipeline for co-speech gestures and multi-party HRI scenarios, contributing more diverse, representative gesture datasets to the gesture community. To our knowledge, this is the first study to systematically demonstrate the feasibility of using generated gestures not only for ML but also for broader cross-disciplinary research. Our pipeline is fully accessible to researchers without immense ML expertise, enabling reproducible gestures across contexts, supporting psychology, robotics, gaming and other fields. We believe this work provides an invaluable foundation to guide future research in gesture generation, recognition and evaluation, and to expand the diversity and representativeness of gesture datasets for downstream tasks.

%%%%%%%%%%%%%%%%%%%%%%%%%%%%%%%%%%%%%%%%%%%%%%%%%%%%%%%%%%%%%%%%%%%%%%%%%%%%%%%%
\section{ACKNOWLEDGMENTS}
We thank Dr. Philipp Allgeuer for his technical expertise with the NICOL robot and Dr. Matthias Kerzel for assisting with participant anonymization in the real dataset.

%%%%%%%%%%%%%%%%%%%%%%%%%%%%%%%%%%%%%%%%%%%%%%%%%%%%%%%%%%%%%%%%%%%%%%%%%%%%%%%%

{\small
\bibliographystyle{ieee}
\bibliography{refs}

@inproceedings{Pozzi22,
  title={{Pointing Gestures for Human-Robot Interaction in Service Robotics: A Feasibility Study}},
  author={Pozzi, Luca and Gandolla, Marta and Roveda, Loris},
  booktitle={International Conference on Computers Helping People with Special Needs},
  pages={461--468},
  year={2022},
  organization={Springer}
}

@article{Jirak21,
  title={{Solving Visual Object Ambiguities when Pointing: An Unsupervised Learning Approach}},
  author={Jirak, Doreen and Biertimpel, David and Kerzel, Matthias and Wermter, Stefan},
  journal={Neural Computing and Applications},
  volume={33},
  number={7},
  pages={2297--2319},
  year={2021},
  publisher={Springer}
}

@inproceedings{Das21,
  title={{A Data-Set and a Method for Pointing Direction Estimation from Depth Images for Human-Robot Interaction and VR Applications}},
  author={Das, Shome S},
  booktitle={2021 IEEE International Conference on Robotics and Automation (ICRA)},
  pages={11485--11491},
  year={2021},
  organization={IEEE}
}

@inproceedings{Huang16,
  title={{A Pointing Gesture Based Egocentric Interaction System: Dataset, Approach and Application}},
  author={Huang, Yichao and Liu, Xiaorui and Zhang, Xin and Jin, Lianwen},
  booktitle={Proceedings of the IEEE conference on computer vision and pattern recognition workshops},
  pages={16--23},
  year={2016}
}

@article{Uraka23,
  title={{Nonverbal Cues in Human--Robot Interaction: A Communication Studies Perspective}},
  author={Urakami, Jacqueline and Seaborn, Katie},
  journal={ACM Transactions on Human-Robot Interaction},
  volume={12},
  number={2},
  pages={1--21},
  year={2023},
  publisher={ACM New York, NY}
}

@article{Peral22,
  title={{Efficient Hand Gesture Recognition for Human-Robot Interaction}},
  author={Peral, Marc and Sanfeliu, Alberto and Garrell, Ana{\'\i}s},
  journal={IEEE Robotics and Automation Letters},
  volume={7},
  number={4},
  pages={10272--10279},
  year={2022},
  publisher={IEEE}
}

@Article{Dick12,
  Title                    = {{Gesture in the Developing Brain}},
  Author                   = {Dick, Anthony Steven and Goldin-Meadow, Susan and Solodkin, Ana and Small, Steven L.},
  Journal                  = {Developmental Science},
  Year                     = {2012},
  Number                   = {2},
  Pages                    = {165-180},
  Volume                   = {15},
  Doi                      = {doi:10.1111/j.1467-7687.2011.01100.x}
}

@ARTICLE{Corba09,
  author = {Corballis, Michael C.},
  title = {{Language as Gesture}},
  journal = {Human Movement Science},
  year = {2009},
  volume = {28},
  pages = {556--565},
  number = {5},
  month = oct,
  comment = {print},
  day = {08},
  doi = {10.1016/j.humov.2009.07.003},
  issn = {01679457},
}

@InProceedings{Kuche23,
  author    = {Kucherenko, Taras and Nagy, Rajmund and Yoon, Youngwoo and Woo, Jieyeon and Nikolov, Teodor and Tsakov, Mihail and Henter, Gustav Eje},
  booktitle = {Proceedings of the 25th International Conference on Multimodal Interaction},
  title     = {{The GENEA Challenge 2023: A Large-Scale Evaluation of Gesture Generation Models in Monadic and Dyadic Settings}},
  year      = {2023},
  pages     = {792--801},
  comment   = {FG26 GENEA challenge, large scale gesture generation},
}

@Article{Kuche24,
  author    = {Kucherenko, Taras and Wolfert, Pieter and Yoon, Youngwoo and Viegas, Carla and Nikolov, Teodor and Tsakov, Mihail and Henter, Gustav Eje},
  journal   = {ACM Transactions on Graphics},
  title     = {{Evaluating Gesture Generation in a Large-Scale Open Challenge: The GENEA Challenge 2022}},
  year      = {2024},
  number    = {3},
  pages     = {1--28},
  volume    = {43},
  comment   = {related work},
  publisher = {ACM New York, NY},
}

@InProceedings{Wang25,
  author       = {Wang, Boyang and Sridhar, Nikhil and Feng, Chao and Van der Merwe, Mark and Fishman, Adam and Fazeli, Nima and Park, Jeong Joon},
  booktitle    = {2025 IEEE International Conference on Robotics and Automation (ICRA)},
  title        = {{This\&That: Language-Gesture Controlled Video Generation for Robot Planning}},
  year         = {2025},
  organization = {IEEE},
  pages        = {12842--12849},
}

@InProceedings{Kim24,
  author    = {Kim, Sunwoo and Chang, Minwook and Kim, Yoonhee and Lee, Jehee},
  booktitle = {SIGGRAPH Asia 2024 Conference Papers},
  title     = {{Body Gesture Generation for Multimodal Conversational Agents}},
  year      = {2024},
  pages     = {1--11},
}

@article{hashi2024,
author = {Hashi, Abdirahman and Mohd Hashim, Siti and Asamah, Azurah},
year = {2024},
month = {01},
pages = {1-1},
title = {{A Systematic Review of Hand Gesture Recognition: An Update From 2018 to 2024}},
volume = {PP},
journal = {IEEE Access},
doi = {10.1109/ACCESS.2024.3421992}
}

@article{khan2025,
author = {Khan, Md Asraful Islam and Sarowar, Md Selim and Islam, Mohaiminul and FARJANA, NUR and ISLAM, SYFUL},
year = {2025},
month = {05},
pages = {33},
title = {{Hand Gesture Recognition Systems: A Review of Methods, Datasets, and Emerging Trends}},
volume = {187},
journal = {International Journal of Computer Applications},
doi = {10.5120/ijca2025924776}
}

@article{foteinos2025,
author = {Foteinos, Konstantinos and Cani, Jorgen and Linardakis, Manousos and Radoglou Grammatikis, Panagiotis and Argyriou, Vasileios and Sarigiannidis, Panagiotis and Varlamis, Iraklis and Papadopoulos, Georgios},
year = {2025},
month = {07},
pages = {},
journal = {arXiv},
title = {{Visual Hand Gesture Recognition with Deep Learning: A Comprehensive Review of Methods, Datasets, Challenges and Future Research Directions}},
volume = {2507.04465},
archivePrefix={arXiv}
}

@article{bao2024,
author = {Bao, Fan and Xiang, Chendong and Yue, Gang and He, Guande and Zhu, Hongzhou and Zheng, Kaiwen and Zhao, Min and Liu, Shilong and Wang, Yaole and Zhu, Jun},
journal = {CoRR},
title = {{Vidu: a Highly Consistent, Dynamic and Skilled Text-to-Video Generator with Diffusion Models}},
year = 2024
}

@ARTICLE{linardakis2025,
  author={Linardakis, Manousos and Varlamis, Iraklis and Papadopoulos, Georgios Th.},
  journal={IEEE Access}, 
  title={{Survey on Hand Gesture Recognition from Visual Input}}, 
  year={2025},
  volume={13},
  number={},
  pages={135373-135406},
  doi={10.1109/ACCESS.2025.3593428}}

@article{zhang2020mediapipe,
title={{MediaPipe Hands: On-Device Real-time Hand Tracking}}, 
author={Fan Zhang and Valentin Bazarevsky and Andrey Vakunov and Andrei Tkachenka and George Sung and Chuo-Ling Chang and Matthias Grundmann},
year={2020},
eprint={2006.10214},
volume={2006.10214},
archivePrefix={arXiv},
journal={arXiv},
primaryClass={cs.CV},
}

@article{moeller2025,
   title={{Pointing-Guided Target Estimation via Transformer-Based Attention}},
   journal={Artificial Neural Networks and Machine Learning – ICANN 2025},
   publisher={Springer},
   author={Luca Möller and Hassan Ali and Philipp Allgeuer and Stefan Wermter},
   year={2025},
}

@inproceedings{patel_2025,
    title={{Robotic Manipulation by Imitating Generated Videos Without Physical Demonstrations}},
    author={Shivansh Patel and Shraddhaa Mohan and Hanlin Mai and Unnat Jain and Svetlana Lazebnik and Yunzhu Li},
    booktitle={Workshop on Foundation Models Meet Embodied Agents at CVPR 2025},
    year={2025},
}

@ARTICLE{kerzel2023_nicol,
  author={Kerzel, Matthias and Allgeuer, Philipp and Strahl, Erik and Frick, Nicolas and Habekost, {Jan-Gerrit} and Eppe, Manfred and Wermter, Stefan},
  journal={IEEE Access}, 
  title={{NICOL: A Neuro-Inspired Collaborative Semi-Humanoid Robot That Bridges Social Interaction and Reliable Manipulation}}, 
  year={2023},
  volume={11},
  number={},
  pages={123531-123542},
  doi={10.1109/ACCESS.2023.3329370}}

@inproceedings{liu2025gesturelsmlatentshortcutbased,
  title={{GestureLSM: Latent Shortcut based Co-Speech Gesture Generation with Spatial-Temporal Modeling}},
  author={Pinxin Liu and Luchuan Song and Junhua Huang and Chenliang Xu},
  booktitle={IEEE/CVF International Conference on Computer Vision},
  year={2025},
}

@inproceedings{heusel2017_fid,
    author = {Heusel, Martin and Ramsauer, Hubert and Unterthiner, Thomas and Nessler, Bernhard and Hochreiter, Sepp},
    title = {{GANs Trained by a Two Time-Scale Update Rule Converge to a Local Nash Equilibrium}},
    year = {2017},
    isbn = {9781510860964},
    publisher = {Curran Associates Inc.},
    address = {Red Hook, NY, USA},
    pages = {6629–6640},
    numpages = {12},
    location = {Long Beach, California, USA},
    series = {NIPS'17}
}

@article{unterthiner_2019_fvd,
title={{Towards Accurate Generative Models of Video: A New Metric \& Challenges}}, 
author={Thomas Unterthiner and Sjoerd van Steenkiste and Karol Kurach and Raphael Marinier and Marcin Michalski and Sylvain Gelly},
year={2019},
eprint={1812.01717},
volume={1812.01717},
archivePrefix={arXiv},
journal={arXiv},
primaryClass={cs.CV},
}

@inproceedings{radford2021_clip,
  title={{Learning Transferable Visual Models from Natural Language Supervision}},
  author={Radford, Alec and Kim, Jong Wook and Hallacy, Chris and Ramesh, Aditya and Goh, Gabriel and Agarwal, Sandhini and Sastry, Girish and others},
  booktitle={International Conference on Machine Learning},
  year={2021},
  organization={PMLR}
}

@Article{qi2024,
author={Qi, Jing
and Ma, Li
and Cui, Zhenchao
and Yu, Yushu},
title={{Computer Vision-Based Hand Gesture Recognition for Human-Robot Interaction: A Review}},
journal={Complex {\&} Intelligent Systems},
year={2024},
month={Feb},
day={01},
volume={10},
number={1},
pages={1581-1606},
doi={10.1007/s40747-023-01173-6},
}

@article{jung_2024_crowdsourcing,
author = {In-Taek Jung and Sooyeon Ahn and JuChan Seo and Jin-Hyuk Hong},
title = {{Exploring the Potentials of Crowdsourcing for Gesture Data Collection}},
journal = {International Journal of Human–Computer Interaction},
volume = {40},
number = {12},
pages = {3112--3121},
year = {2024},
publisher = {Taylor \& Francis},
doi = {10.1080/10447318.2023.2180235},
eprint = {https://doi.org/10.1080/10447318.2023.2180235}
}

@article{gao_2024_challenges,
title = {{Challenges and Solutions for Vision-Based Hand Gesture Interpretation: A Review}},
journal = {Computer Vision and Image Understanding},
volume = {248},
pages = {104095},
year = {2024},
issn = {1077-3142},
doi = {https://doi.org/10.1016/j.cviu.2024.104095},
author = {Kun Gao and Haoyang Zhang and Xiaolong Liu and Xinyi Wang and Liang Xie and Bowen Ji and Ye Yan and Erwei Yin},
}

@article{liu_2024_sora_review,
title={{Sora: A Review on Background, Technology, Limitations, and Opportunities of Large Vision Models}}, 
author={Yixin Liu and Kai Zhang and Yuan Li and Zhiling Yan and Chujie Gao and Ruoxi Chen and Zhengqing Yuan and Yue Huang and Hanchi Sun and Jianfeng Gao and Lifang He and Lichao Sun},
year={2024},
eprint={2402.17177},
volume={2402.17177},
archivePrefix={arXiv},
journal={arXiv},
primaryClass={cs.CV},

}

@inproceedings{dai_2024_safesora,
title={{SafeSora: Towards Safety Alignment of Text2Video Generation via a Human Preference Dataset}},
author={Josef Dai and Tianle Chen and Xuyao Wang and Ziran Yang and Taiye Chen and Jiaming Ji and Yaodong Yang},
booktitle={The Thirty-eight Conference on Neural Information Processing Systems Datasets and Benchmarks Track},
year={2024}
}

@article{li_2025_zerohsi,
journal = {arXiv},
title={{ZeroHSI: Zero-Shot 4D Human-Scene Interaction by Video Generation}}, 
author={Hongjie Li and Hong-Xing Yu and Jiaman Li and Jiajun Wu},
year={2025},
volume={2412.18600},
archivePrefix={arXiv},
primaryClass={cs.CV},

}

@inproceedings{deichler_2023,
author = {Deichler, Anna and Mehta, Shivam and Alexanderson, Simon and Beskow, Jonas},
title = {{Diffusion-Based Co-Speech Gesture Generation Using Joint Text and Audio Representation}},
year = {2023},
isbn = {9798400700552},
publisher = {Association for Computing Machinery},
address = {New York, NY, USA},
doi = {10.1145/3577190.3616117},
booktitle = {Proceedings of the 25th International Conference on Multimodal Interaction},
pages = {755–762},
numpages = {8},
location = {Paris, France},
series = {ICMI '23}
}

@inproceedings{zhao_2023,
author = {Zhao, Weiyu and Hu, Liangxiao and Zhang, Shengping},
title = {{DiffuGesture: Generating Human Gesture From Two-person Dialogue With Diffusion Models}},
year = {2023},
isbn = {9798400703218},
publisher = {Association for Computing Machinery},
address = {New York, NY, USA},
doi = {10.1145/3610661.3616552},
booktitle = {Companion Publication of the 25th International Conference on Multimodal Interaction},
pages = {179–185},
numpages = {7},
location = {Paris, France},
series = {ICMI '23 Companion}
}

@InProceedings{liu_2024_emage,
author    = {Liu, Haiyang and Zhu, Zihao and Becherini, Giorgio and Peng, Yichen and Su, Mingyang and Zhou, You and Zhe, Xuefei and Iwamoto, Naoya and Zheng, Bo and Black, Michael J.},
title     = {{EMAGE: Towards Unified Holistic Co-Speech Gesture Generation via Expressive Masked Audio Gesture Modeling}},
booktitle = {Proceedings of the IEEE/CVF Conference on Computer Vision and Pattern Recognition (CVPR)},
month     = {June},
year      = {2024},
pages     = {1144-1154}
}

@inproceedings{singer_2023_makeavideo,
title={{Make-A-Video: Text-to-Video Generation without Text-Video Data}},
author={Uriel Singer and Adam Polyak and Thomas Hayes and Xi Yin and Jie An and Songyang Zhang and Qiyuan Hu and Harry Yang and Oron Ashual and Oran Gafni and Devi Parikh and Sonal Gupta and Yaniv Taigman},
booktitle={The Eleventh International Conference on Learning Representations },
year={2023},
}

@article{ho_2022_imagenvideo,
title={{Imagen Video: High Definition Video Generation with Diffusion Models}}, 
author={Jonathan Ho and William Chan and Chitwan Saharia and Jay Whang and Ruiqi Gao and Alexey Gritsenko and Diederik P. Kingma and Ben Poole and Mohammad Norouzi and David J. Fleet and Tim Salimans},
year={2022},
eprint={2210.02303},
volume={2210.02303},
archivePrefix={arXiv},
primaryClass={cs.CV},
journal = {arXiv},
}

@INPROCEEDINGS{wu_2023_tuneavideo,
author={Wu, Jay Zhangjie and Ge, Yixiao and Wang, Xintao and Lei, Stan Weixian and Gu, Yuchao and Shi, Yufei and Hsu, Wynne and Shan, Ying and Qie, Xiaohu and Shou, Mike Zheng},
booktitle={2023 IEEE/CVF International Conference on Computer Vision (ICCV)}, 
title={{Tune-A-Video: One-Shot Tuning of Image Diffusion Models for Text-to-Video Generation}}, 
year={2023},
volume={},
number={},
pages={7589-7599},
doi={10.1109/ICCV51070.2023.00701}}

@INPROCEEDINGS{khachatryan_2023_text2videozero,
author={Khachatryan, Levon and Movsisyan, Andranik and Tadevosyan, Vahram and Henschel, Roberto and Wang, Zhangyang and Navasardyan, Shant and Shi, Humphrey},
booktitle={2023 IEEE/CVF International Conference on Computer Vision (ICCV)}, 
title={{Text2Video-Zero: Text-to-Image Diffusion Models are Zero-Shot Video Generators}}, 
year={2023},
volume={},
number={},
pages={15908-15918},
doi={10.1109/ICCV51070.2023.01462}
}

@InProceedings{zimmermann_2019_freihand,
author    = {Christian Zimmermann and Duygu Ceylan and Jimei Yang and Bryan Russell and Max Argus and Thomas Brox},
title     = {{FreiHAND: A Dataset for Markerless Capture of Hand Pose and Shape from Single RGB Images}},
booktitle = {IEEE International Conference on Computer Vision (ICCV)},
year = {2019},
}

@article{li_2023_renderih,
  title={{RenderIH: A Large-scale Synthetic Dataset for 3D Interacting Hand Pose Estimation}},
  author={Lijun Li and Linrui Tian and Xindi Zhang and Qi Wang and Bang Zhang and Liefeng Bo and Mengyuan Liu and Chen Chen},
  journal={2023 IEEE/CVF International Conference on Computer Vision (ICCV)},
  year={2023},
  pages={20338-20348},
}

@InProceedings{kwon_2021_h2o,
author = {Kwon, Taein and Tekin, Bugra and St\"uhmer, Jan and Bogo, Federica and Pollefeys, Marc},
title = {{H2O: Two Hands Manipulating Objects for First Person Interaction Recognition}},
booktitle = {Proceedings of the IEEE/CVF International Conference on Computer Vision (ICCV)},
month = {October},
year = {2021},
pages = {10138-10148}
}

@ARTICLE{wan_2022_chalearn,
  author={Wan, Jun and Lin, Chi and Wen, Longyin and Li, Yunan and Miao, Qiguang and Escalera, Sergio and Anbarjafari, Gholamreza and Guyon, Isabelle and Guo, Guodong and Li, Stan Z.},
  journal={IEEE Transactions on Cybernetics}, 
  title={{ChaLearn Looking at People: IsoGD and ConGD Large-Scale RGB-D Gesture Recognition}}, 
  year={2022},
  volume={52},
  number={5},
  pages={3422-3433},
  doi={10.1109/TCYB.2020.3012092}}

@misc{sora_2024_videoworldsimulators2024,
  title={{Video Generation Models as World simulators}},
  author={Tim Brooks and Bill Peebles and Connor Holmes and Will DePue and Yufei Guo and Li Jing and David Schnurr and Joe Taylor and Troy Luhman and Eric Luhman and Clarence Ng and Ricky Wang and Aditya Ramesh},
  year={2024}
}

@misc{veo_2024,
  title={{Veo: Generative Video Model}},
  author={{Google DeepMind}},
  year={2024}
}

@ARTICLE{hai_2025_sora_limitation,    
AUTHOR={Hai, Tran Trieu  and Mai, Duong Thi Thuy  and Hanh, Nguyen Van },          
TITLE={{A Rapid Review of Using AI-Generated Instructional Videos in Higher Education}},    
JOURNAL={Frontiers in Computer Science},     
VOLUME={Volume 7 - 2025},
YEAR={2026},
URL={https://www.frontiersin.org/journals/computer-science/articles/10.3389/fcomp.2025.1721093},
DOI={10.3389/fcomp.2025.1721093},
ISSN={2624-9898},
}

@article{motamed_2025_generativevideomodelsunderstand,
      title={{Do Generative Video Models Understand Physical Principles?}}, 
      author={Saman Motamed and Laura Culp and Kevin Swersky and Priyank Jaini and Robert Geirhos},
      year={2025},
      volume = {arXiv},
      eprint={2501.09038},
      archivePrefix={arXiv},
      journal = {arXiv},
}

@inproceedings{tong_2022_videomae,
author = {Tong, Zhan and Song, Yibing and Wang, Jue and Wang, Limin},
title = {{VideoMAE: Masked Autoencoders are Data-Efficient Learners for Self-Supervised Video Pre-Training}},
year = {2022},
isbn = {9781713871088},
publisher = {Curran Associates Inc.},
address = {Red Hook, NY, USA},
booktitle = {Proceedings of the 36th International Conference on Neural Information Processing Systems},
articleno = {732},
numpages = {16},
location = {New Orleans, LA, USA},
series = {NIPS '22}
}
}

\end{document}